\documentclass{article}

 \usepackage[final,nonatbib]{nips_2018}

\usepackage[utf8]{inputenc} 
\usepackage[T1]{fontenc}    
\usepackage{hyperref}       
\usepackage{url}  
\usepackage{booktabs}       
\usepackage{amsfonts}       
\usepackage{nicefrac}       
\usepackage{microtype}      
\usepackage{shortcuts}
\usepackage{graphicx}
\usepackage{subfig}
\usepackage{algorithm,algorithmic}
\usepackage{amsmath}
\graphicspath{{figs/}}
\makeatletter
\g@addto@macro\normalsize{%
  \setlength\abovedisplayskip{-1pt}
  \setlength\belowdisplayskip{-1pt}
  \setlength\abovedisplayshortskip{-1pt}
  \setlength\belowdisplayshortskip{-1pt}
}
\makeatother
\usepackage[font=small,skip=5pt]{caption}
 \pdfoutput=1
\title{Personalizing Intervention Probabilities By Pooling} 

\author{
Sabina Tomkins$^{*1}$  \quad Predrag Klasnja$^\dagger$  \quad Susan Murphy$^{*2}$\\
Harvard University$^*$   \quad University of Michigan$^\dagger$\\
\texttt{\{sabina\_tomkins$^{1}$,samurphy$^{2}$\}@fas.harvard.edu} \quad \texttt{klasnja@umich.edu}$^\dagger$ 
  }

\begin{document}

\maketitle

\begin{abstract}
In many mobile health interventions,  treatments should only be delivered
in a particular context, for example when a user is currently stressed, walking or sedentary. 
 Even in an optimal context, 
 concerns about user burden can restrict which treatments are sent. 
To diffuse the treatment delivery 
over times 
when a user is in a desired context, 
it is critical to predict the future number of times the context will occur.
The focus of this paper is on whether  personalization can  improve predictions in these settings. 
Though 
the variance between 
individuals' behavioral patterns 
suggest that personalization should be useful, 
the amount of individual-level data limits its capabilities.
Thus, we investigate several methods which pool data across users to overcome these deficiencies 
and
find that 
pooling 
lowers the overall error rate relative to both personalized and batch approaches. 

\commentout{
In mobile health settings it is critical to know when to deploy an intervention. 
The decision of whether to intervene can be informed by knowledge of when certain events might occur. 
For example, when determining whether to send a message to increase activity, 
it is helpful to know how many more times a participant might be inactive or sedentary. 
Personalization can improve predictions in these settings, as 
individuals'  behavioral patterns  
vary greatly. 
However, the amount of individual-level data limits the capabilities of full personalization and
we investigate several methods which utilize levels of pooling to overcome these deficiencies. 
We
find that pooling 
achieves the best overall error rate compared to both personalized and batch approaches. 
Yet, the effectiveness of pooling is influenced by the stationarity of individuals' behavior. 
}
\commentout{
As individuals' activity patterns can 
vary greatly, and irrelevant interventions can drive disengagement,
we propose to personalize these predictions.

Our goal is to predict
the number of sedentary periods which remain in a day.
This information 
can 
then be used to inform the choice of sending an intervention at a given time.
As individuals' activity patterns can 
vary greatly, and irrelevant interventions can drive disengagement,
we propose to personalize these predictions.
However, as the amount of individual data limits the capabilities of full personalization 
we investigate several methods which utilize levels of pooling to overcome these deficiencies. 
We
show that pooling 
achieves the best overall error rate compared to both personalized and batch approaches. 
}

\end{abstract}

\setlength{\abovedisplayskip}{0pt}
\setlength{\belowdisplayskip}{0pt}

\section{Introduction}

Mobile health (mHealth) interventions can deliver effective treatments in real-time. 
For example, to help users increase their physical activity, an mHealth application
might send a suggestion to walk at a time when a user is motivated and able 
to pursue the suggestion. Users struggling to quit smoking might be prompted to do a mindfulness 
exercise when the system detects they are becoming stressed, helping reduce negative affect and the 
desire to smoke. The promise of mHealth interventions hinges on their ability to provide support at times when users need the support and are receptive to it \cite{nahum2017}.

Many interventions included in an mHealth system (e.g. reminders, coping strategies) are designed to be delivered in a particular context.   These treatments are often delivered via a wearable or a push notification on a smartphone.
An approach to reducing user burden is to budget the number of treatments delivered in a certain period of time (e.g., a day). 
Given an intervention budget, our goal is to spread this budget across
 the instances in which a user is in a targeted context (e.g., currently sedentary).
 By doing so we aim 
 to 
  improve user experience.
 
 
 

Consider an mHealth application designed to reduce sedentary behavior. The application can send users an activity suggestion when they have been sitting for 40 minutes or longer, but the number of suggestions sent each day is constrained by a budget. The subject of this paper is predicting the number of future sedentary episodes remaining in the day. This prediction is used to decide whether to send a suggestion. For example, it would be more important to send a suggestion now if the system predicted only a single additional sedentary episode than if many were likely. 
As this sedentary behavior can have high variance from person to person, we argue for a personalized approach. 

\commentout{


If only a single sedentary episode was predicted in the remainder of the day, and the budget was not yet met, it would make sense to send a reminder. Alternatively, if many future sedentary episodes were predicted, it is less important to send a reminder now. 

Mobile health interventions engage with participants to achieve desired health outcomes. 
Consider the healthcare goal of lowering risk of heart-related illnesses. A target health outcome might be 
to increase physical activity among a population of participants, measured in daily step counts. A mobile health intervention might be 
to send reminders to be active, or to notify participants of their current step count. 
Examples in mobile health abound, from lowering risk of stress by sending mindfulness activity suggestions, 
to facilitating rehabilitation by reminding participants of their medication schedules.

Across mobile health settings a persistent problem is engaging with participants
without over-burdening them. A response to this problem is to impose a budget on the
number of messages sent within a period of time, for example, a budget might be
2 messages per day. 
Given a budget of  interactions, our goal is to spread this budget across a period of time. 
Doing so is instrumental in learning the contexts under which users are receptive to reminders. As we collect data on the interaction between context and receptivity, we can better adapt interventions to be meaningful to each user. 

The subject of this paper is  predicting the number of future sedentary episodes in the remainder of the day. 

Given a budget of  interactions, our goal is to spread this budget across all remaining sedentary periods. Doing so is instrumental in learning the contexts under which users are receptive to reminders. As we collect data on the interaction between context and receptivity, we can better adapt interventions to be meaningful to each user.

To reduce burden, we might have a budget for the total number of reminders sent in a day. 
If only a single sedentary episode was predicted in the remainder of the day, and the budget was not yet met, it would make sense to send a reminder. Alternatively, if many future sedentary episodes were predicted, it is less important to send a reminder now. 

The subject of this paper is  predicting the number of future sedentary episodes in the remainder of the day. As this quantity can have high variance from person to person, we argue for a personalized approach. Given a budget of daily activity reminders, our goal is to spread this budget across all remaining sedentary periods. Doing so is instrumental in learning the contexts under which users are receptive to reminders. As we collect data on the interaction between context and receptivity, we can better adapt interventions to be meaningful to each user.

 is characterizing effective 
interactions with participants, without not over-burdening them. 
For example, people may be more responsive to different messages at different times, 
and we would like to learn which messages are most effective to whom, when. 
However,

 how often to interact with participants without over-burdening them.

While a sufficient level of interaction must be maintained in order to engage participants, interacting too often can place a 
over-burden on users. 

To successfully engage participants, a sufficie

In mobile health settings, a common type of intervention is a message designed to achieve a health outcome.
For example, weight loss might be a desired health outcome, 
and mobile health application might send 
 reminders to walk or push notifications of one's current activity levels.

Upon enrolling in a study, or downloading a mobile health application, participants/users 
consent to receive many such messages. Even when one is committed to achieving a 
health outcome, these messages can quickly become burdensome.

 In a health study, participants will receive many such messages. 

 a common goal  is to d  designed to achieve a health outcome. For example,  one

send participants/users content throughout the day.

For example, this content might be an encouragement or reminder to do an activity, an uplifting message, or a request for information which will be useful in designing interventions. Participants can become quickly burdened by these messages and an open question is determining how to interact with users while minimizing burden. Consider the example of a notification meant to encourage walking. If a model had knowledge of the number of future sedentary episodes in the day, it might randomize future reminders using this knowledge. 
To reduce burden, we might have a budget for the total number of reminders sent in a day. 

}

 Personalization has been a recent focus in healthcare prediction tasks \cite{shoeb2004,rudovic2018}. 
 Lopez and Picard \cite{lopez2017} proposed a multi-task neural network for pain detection and Saeed and Trajanovski \cite{saeed2017} proposed a multi-task neural network approach to detect stress. 
Here, we consider two forms of pooling. We demonstrate that these approaches allow us to more quickly make appropriate personalized predictions than a fully personalized approach which requires more days of data. Furthermore, pooling approaches outperform a batch model. 




\section{Models}
We predict the number of sedentary times remaining in a day ($t$), for a person $i$. 
Here, we propose two approaches, a Gaussian Process \gp{} and a population-informed task-specific model we refer to as \weighted{} ({}\smallw). 
These are compared to task-specific and batch models.


We treat each person as a task, such that we have $M$ tasks for a population of size $M$. Across all models we consider a dataset of inputs, $X = x^1,\dots,x^N$, and outputs $\mathbf{y} = y^1,\dots,y^N$ where each $x^i$ belongs to a task in $\{1..M\}$, and some $y's$ are given while others are unknown.
Given a set of indices which define a test set, $\mathrm{T}\subseteq \{1..M\}$,  we are interested in inferring $\hat{y}^i$ for each $i$ in $T$. 

\subsection{Gaussian Process Variants}
A natural question in this setting is how well a task-specific \gp{} would perform. Thus we introduce, \gpind{} which learns some set of parameters separately for each person. We compare this to \gpbatch{} which treats all input as belonging to one task, and learns parameters for this population-size task. 
Both \gpind{} and \gpbatch{} assume standard \gp forms, that is a prediction $y^i$ is made as:

 $$y^i \sim \mathcal{N}\big((K(x^i,X), K(X,X) )^{-1}\mathbf{y}, K(x^i,x^i)-K(x^i,X)K(X,X)^{-1}K(X,x^i) \big),$$

where in \gpind{}, $X$ is the training data for one task  $\in T$, and in \gpbatch{} $X$ is the training data for all tasks $\notin T$. 
Additionally, we introduce  \gpmt{}. This model learns some individualized parameter values, as well as some which are shared across the population.  For example, we explicitly learn a $M \times M$ inter-task similarity matrix $K^f$.
\commentout{
$\Theta^i$ for each person $i$. Additionally, we are interested in how pooling might help, and we introduce,  \gpmt{}, which, for person $i$, learns $\Theta$ which can be represented with a subset $\theta^i \in \Theta$, unique to the individual and a subset $\theta^b \in \Theta$, which contains population information.
 Finally, we compare each of these models to \gpbatch{}, which learns $\Theta$ across the entire population.} 
For example, in the multi-task case, a prediction for person(task) $i$,  is made according to: 
 
 $$y^i  = (k_f^i \otimes k_*^x)^T\Sigma^{-1}\mathbf{y} \hspace{4mm} \Sigma = K^f \otimes K^x + D\otimes I,$$

where, $\otimes$ denotes the Kronecker product, $k_l^f $selects the $l^{th}$ column of $K^f$, $k_*^f$,
is the vector of
covariances between the test point $x^i$ and the training points, $K^x$
is the matrix of covariances
between all pairs of training points, $D$ is an $M \times M $diagonal matrix in which the $(l, l)^{th}$
 element is $\sigma^2_l$, and $\Sigma$ is an $MN \times MN$ matrix \cite{bonilla2008}.  
For each model we have use a Radial Basis Function (RBF) kernel for the variance, and a constant mean kernel for the mean.  We place a Gaussian prior on the mean function, where this prior has constant mean and the scale is a tuneable hyper-parameter.

\subsection{Weighted Regressors}
A different approach might be to learn the importance of a batch model relative to a task-specific model.
 To do so we introduce a simple approach which we refer to as weighted regressors. Here, we learn one population level regressor and one individual regressor.
The final prediction is the weighted average of the predictions of these two models, where we adjust the population and task-specific weights, $\beta_g$ and $\beta_l$ respectively, in order to shift weight towards the task-specific model as we gain additional information. This is shown in \algref{wr}. 

\commentout{
 For a training vector $x$, we input x into each regressor and obtain $y^{p*}$, and $y^{d*}$ from the population and individual models respectively. These output form a new input $<y^{p*},y^{i*}>$ and together with the truth $y^i$ we train a linear regressor which can learn to weight each model accordingly. The final output is then $\hat{y}^{i}_t = w_p\times y^{p*}+w_i\times y^{i*}$. We show this procedure in \algref{wr}, here we adopt a sliding window approach where we train on past data up to our current time and predict a few days out. 
}

\begin{algorithm}
\caption{\weighted{} (\smallw)\label{wr}}
\begin{algorithmic}[1]
  \scriptsize
  \STATE{ INPUT:  \quad dataset $\{(x^1,y^1),\dots,(x^M,y^M)\}$,\quad test \,$\mathrm{T}\subseteq \{1..M\}$,\quad window size $w\in[1,n]$,\quad initial training days $ s \in[1,n] $}
    \STATE{ OUTPUT:\quad $\mathbf{\hat{y}} = \{y^j : j\in T\}$}
  \STATE {$R^g = R.\mathrm{train}\big( \{(x^i,y^i) : i \notin T\}\big) $} \hspace{4mm} R \text{refers to some class of regressors}
 \FOR {$ j \in  T$}
    \STATE {$\text{Initialize}: \beta_l,\beta_g = \frac{1}{2}, \delta =.2\times(\mathrm{len}(x^j)/w)^{-1}, \hspace{4mm}$}
  \FOR{ $t \ =\ s,\, s\!+\!w,\, s\!+\!2w,\dots\quad t<\mathrm{len}(x^j)$ }
    \STATE {$(x^{j}_{t} ,y^{j}_{t}) = x_{t+ws}^j , y_{t+ws}^j $}
  \STATE  {$R_l = R.\mathrm{train}\big(\{(x_0^j,y_0^j),\dots,(x_t^j,y_t^j)\}\big)$}     
  \STATE {$y^{g} = R^g(\{x_t^j,\dots,x_{t+w}^j\} ), y^{l} = R^l(\{x_t^j,\dots,x_{t+w}^j\})$  \text{we predict a window length into the future}}
  \STATE   {$\hat{y}^{j}_{t:t+ws} = \beta_gy^{g} + \beta_ly^{l} $}
   \STATE   {$ \beta_g=\beta_g - \delta, \beta_l =\beta_l + \delta$}
 \ENDFOR
 \ENDFOR
\end{algorithmic}
\end{algorithm}

\commentout{
$$p(y_*^i )  \sim N(y_*^i | \mu_* , K_{y_*^i,y_*^i})$$ , 

$$\mu_* = K^{mt}_{f_*,f}(K^{mt}_{f,f}+\Sigma)^{-1}\bf{y}$$
$$K_{y_*^i,y_*^i}= K^{mt}_{f_*,f}(K^{mt}_{f,f}+\Sigma)^{-1}\bf{y}$$
}
\section{Empirical Evaluation} 
It is critical that we obtain high quality predictions in a short amount of time. Thus, to evaluate our methods we consider the setting where we obtain training data on a regular basis and we would like to forecast several days into the future. 
We use data obtained from a real-life mobile health study \cite{klasnja2018}. The goal of the study is to positively impact participants' long-term health by 
increasing their physical activity.  
Participants receive messages which remind them to be active, and step counts are collected as a measure of both overall activity and responsivity to messages. We can incorporate knowledge of predicted sedentary periods in  determining whether to send a message at a certain time. We  label an interval of 40 minutes as sedentary if the total step count in this period is less than 140. 



For simplicity, we predict the number of sedentary periods between 15:00 and 21:00 on each day. In order to use daily context, we focus only on those days with some step count data in the time of 9:00 to 15:00, as well as between 15:00 and 21:00. Additionally, we restrict our attention to those users with at least a week of data. This results in a dataset of 36 participants with approximately 30 days of usable data per participant. For each model and participant we make sliding window predictions where we train with all available data and predict some number of days into the future. 

To capture context we model users' past activity levels and external context. For example, each $x^i_t$ contains the step count of the morning of $t$, the overall step count at $t-1$ and the number of sedentary periods at $t-1$. Additionally, each $x^i_t$ contains a single weather description. These are obtained from \footnote{\url{https://www.kaggle.com/selfishgene/historical-hourly-weather-data}}, where weather is described with a short text description, such as \textit{partly cloudy}. In total there are 21 weather descriptions and the description for  $x^i_t$ is a categorical variable.

\subsection{Sources of variance}
We expect that the extent to which pooling will be advantageous depends on the sources of variance in this data. For example, if the individuals' behavior across time varies greatly it will be difficult to train a personalized model, and we expect some pooling to provide better performance. However, if  behavior varies too greatly from person to person we might see the shared models perform poorly. Thus, before turning to the results we briefly inspect these two sources of variance. 

\textbf{Individual's behavior across time}
To inspect within-participant variance, we perform a $\chi^2$ log-likelihood difference test. For each participant, we set a window length which intuitively corresponds 
to the number of days over which a person might have stable behavior.
For example, if there was no overall temporal patterns, such that each day could be treated as an independent observation we might train a model on each day's data. However, if people behave relatively stably over a period of two days, we could train a new model every two days. 


For a given window length we have 
 $k$ models, where $k $ equals the  
 number of days the participant was in the study divided by the window length. We then train $k$ Ordinary Least Squares regressors. 
For each participant, we create a dataset of roughly 30 days, $(x_i,y_i),\dots,(x_{30},y_{30})$, where each $x_i$ is a vector with context from day $i$ and day $i-1$, and each $y_i$ is the number of sedentary periods from 15:00-21:00 on day $i$. 
To perform the log-likelihood difference test we form a null model which learns one set of parameters over the entirety of a participant's history. This is compared to the aggregate of the $k$ window-length models. Thus, for each participant we determine that it would be better to train $k$ models if:
\begin{center}$\small{-2 \big(\sum_{i=1}^k \log(\mathcal{L}(\mathcal{M}_0))- log(\mathcal{L}(m_i) )\big) > \chi^2_{k-1}},$\end{center}

where $\mathcal{M}_0$ is the baseline model which uses all of the historical data, and each $m_i$ is the model for the $i^{th}$ window. In \figref{ns} we see that when considering only a single day of data at a time most participants change from day to day. However, at windows of length five we see almost no loss in stationarity relative to training one model of window length 30. This might direct the choice to pool or not, where pooling might offer more advantages for smaller window lengths.


\textbf{Similarity of behavior between participants}
To assess between-participant similarities we perform dynamic time warping (DTW) between all pairs of participants. \figref{dtw} shows  
similarities between users. We see both large swaths of dissimilarity (light yellow regions) and pockets of high similarity. This provides some optimism for pooling, while also highlighting the need for personalization.

\begin{figure}[!htb]
    \centering
    \subfloat[Fitting a model every five days is roughly equivalent to fitting one for all days.  \label{ns}]{{\includegraphics[width=0.28\linewidth]{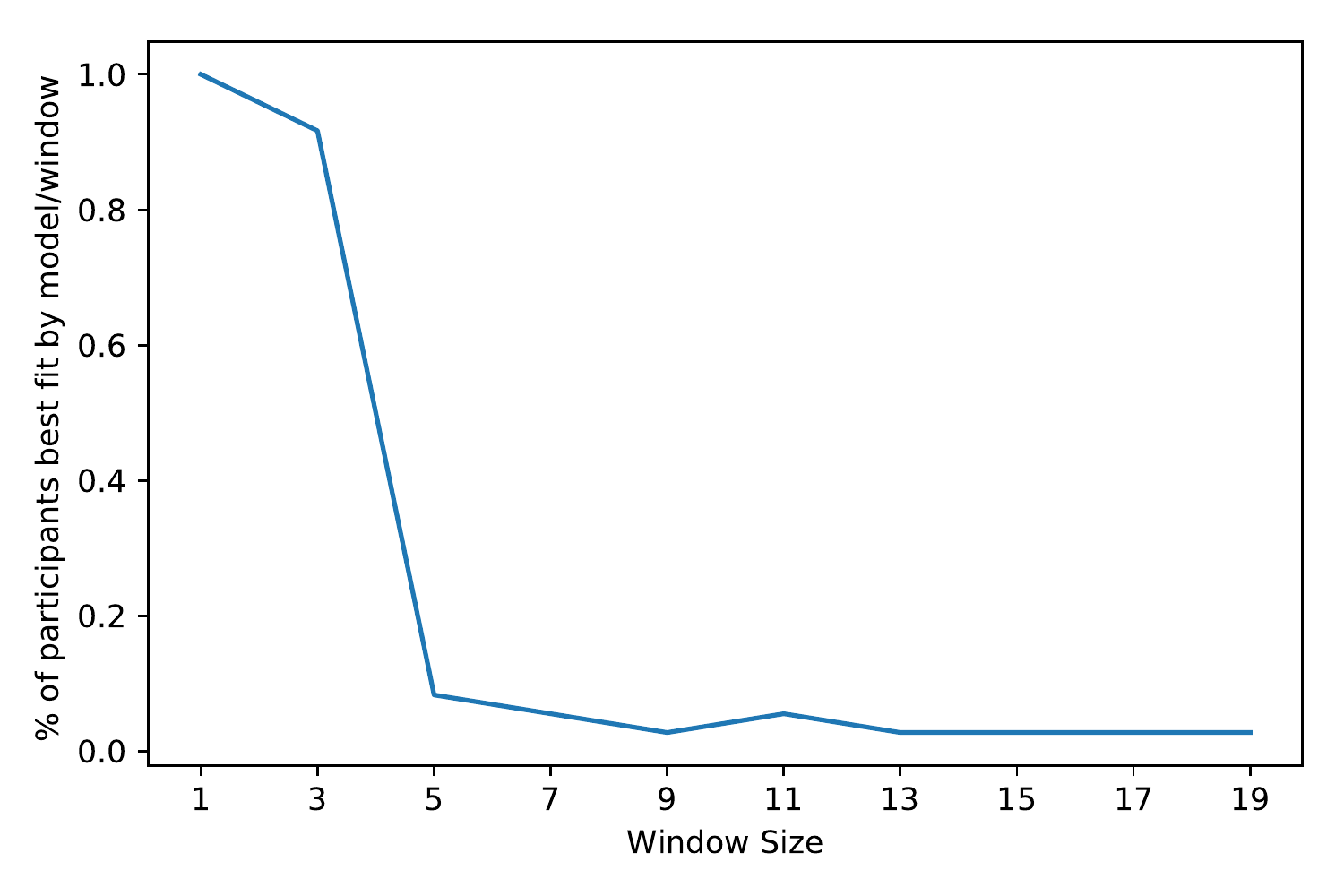} }}%
    \qquad
        \centering
    \subfloat[Similarity as inverse DTW score \label{dtw}]{{\includegraphics[width=0.33\linewidth,trim=.5cm .5cm .5cm .5cm,clip]{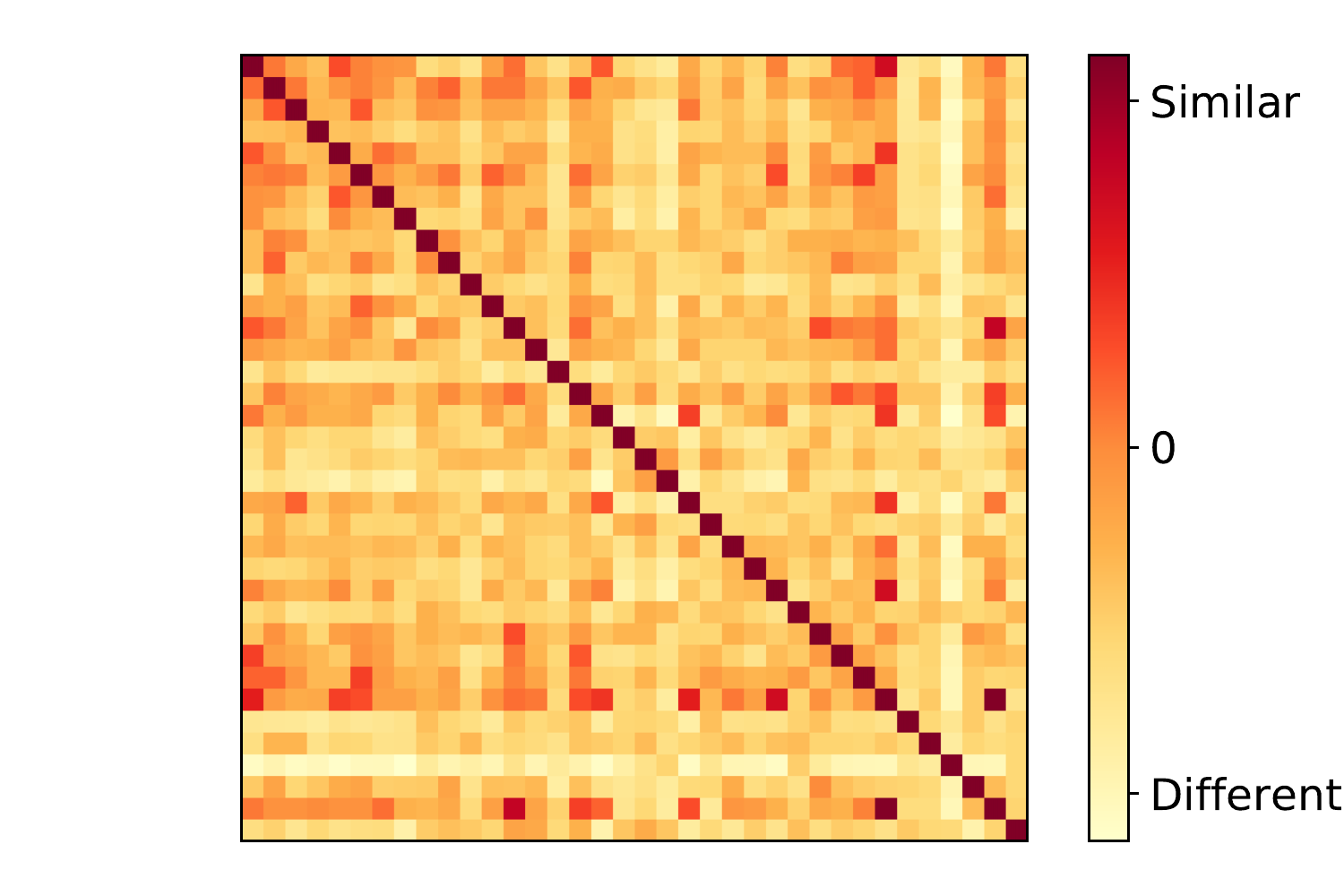} }}%
    \caption{Within-person variance and potentials for pooling.}%
\end{figure}

\commentout{
\begin{figure}[!htb]
    \centering
    \begin{minipage}{.4\textwidth}
        \includegraphics[width=0.9\linewidth]{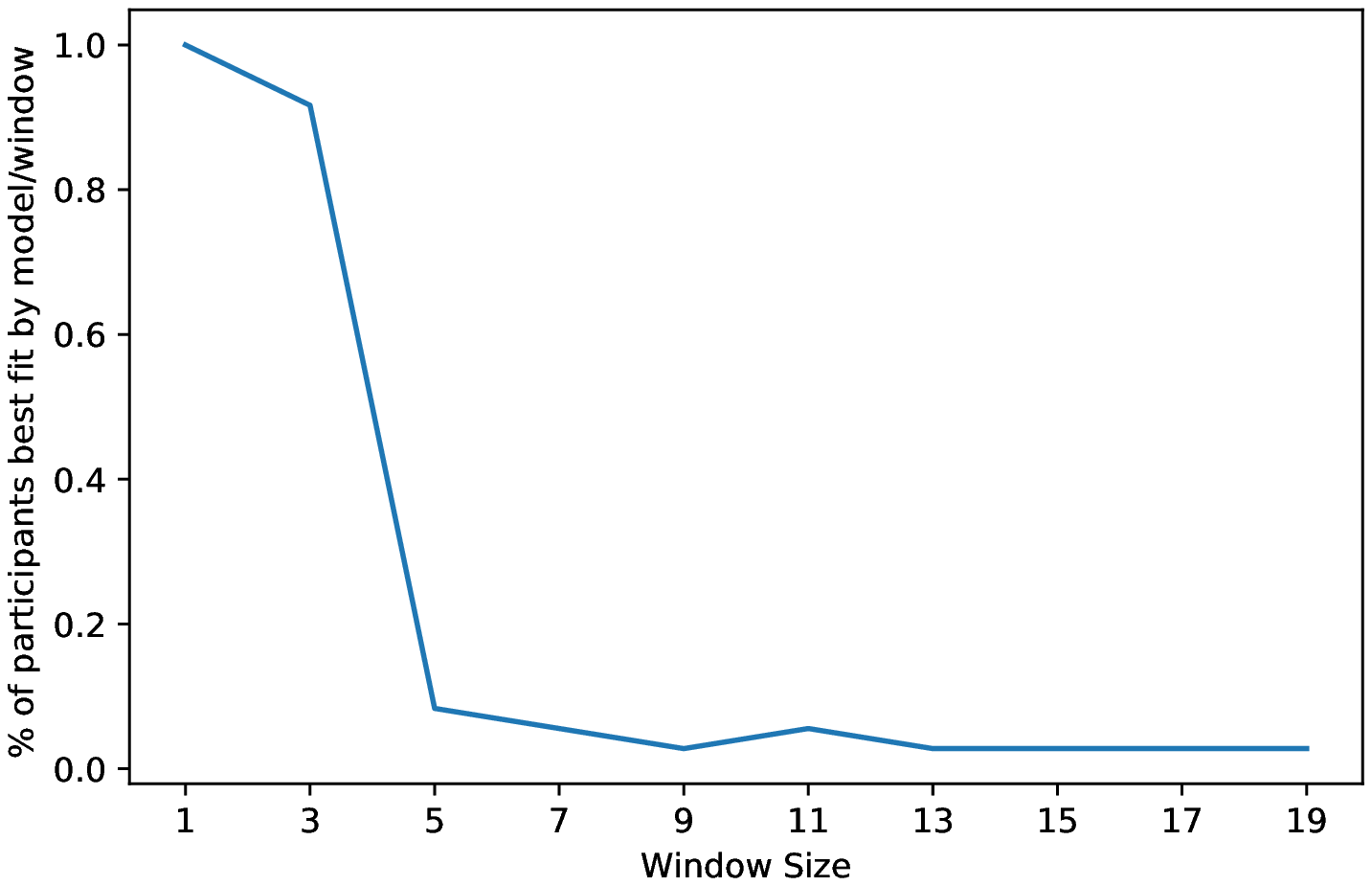}
        \caption{Considering periods of five days, most participants have low time variance. \label{ns}}
    \end{minipage}%
    \begin{minipage}{.4\textwidth}
        \includegraphics[width=0.9\linewidth]{dtw_sims}
        \caption{Similarity as inverse DTW score. \label{dtw}}
    \end{minipage}%
\end{figure}
}


\subsection{Experimental Results}
\label{results}
\begin{figure}[!htb]
    \centering
    \begin{minipage}{.33\textwidth}
        \includegraphics[width=0.9\linewidth]{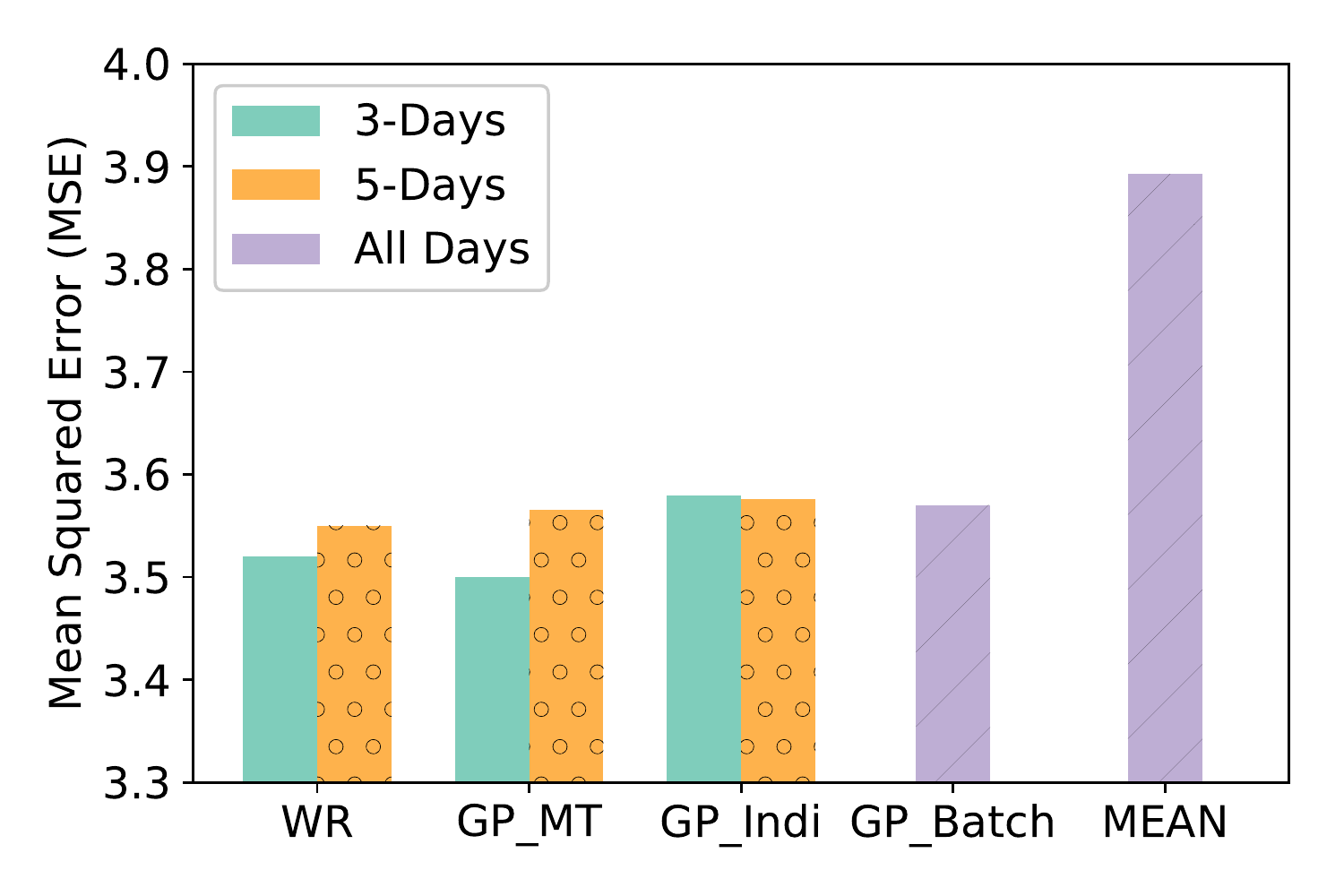}
        \caption{Average error on predicting sedentary periods. \label{mse}}
    \end{minipage}%
    \begin{minipage}{0.33\textwidth}
        \includegraphics[width=0.9\linewidth]{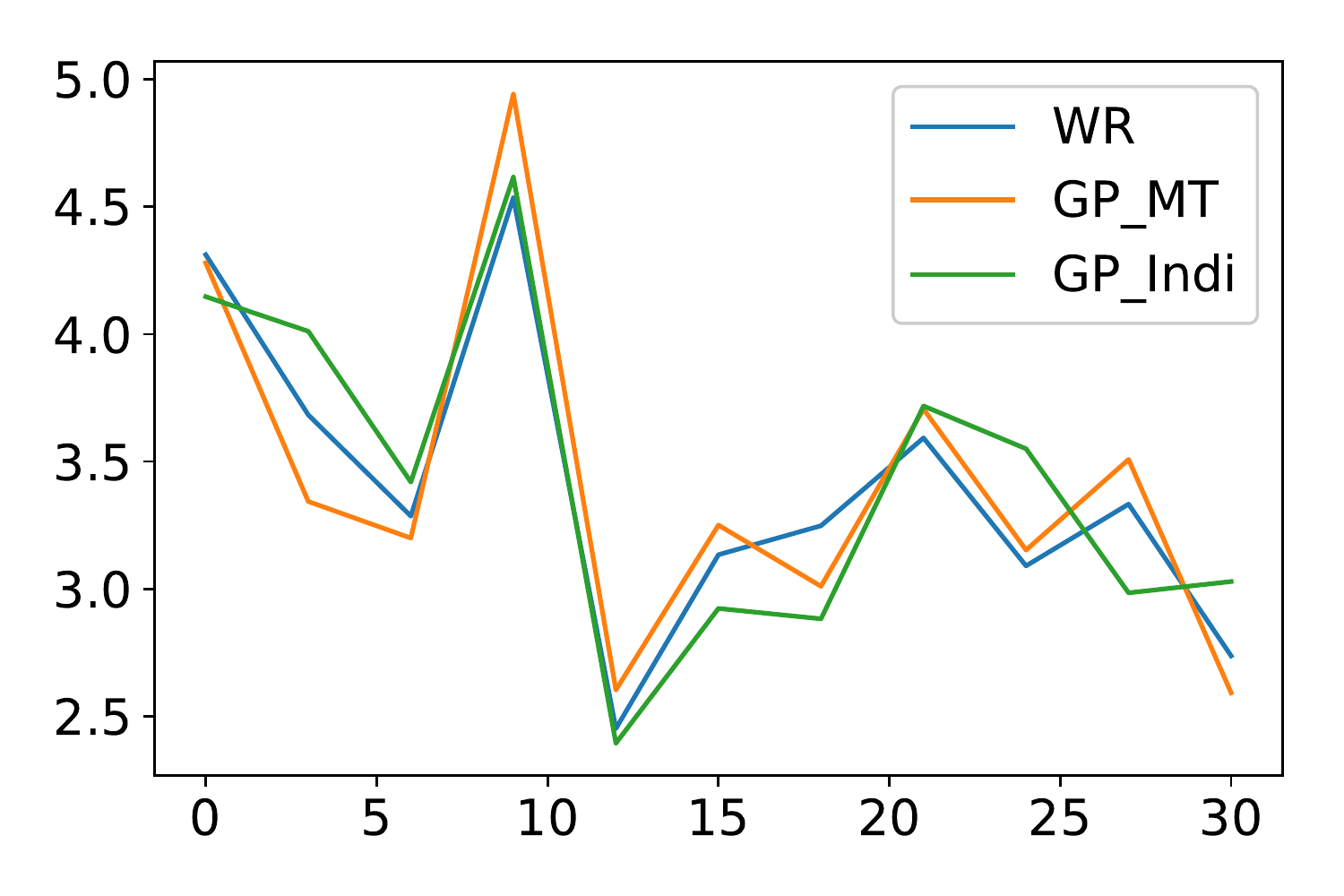}
        \caption{Error over day in study.\label{delta}}
          \hspace{0.6mm}
    \end{minipage}
\end{figure}

We assess the overall Mean Squared Error (MSE)(in \figref{mse}), and evaluate how this changes with increasing amounts of training data in \figref{delta}. 
All results are obtained from five-fold cross validation. We split according to participants, 
as this mirrors the situation where a new participant joins a study and data from previous participants will be utilized for their predictions.
The \gp{} models are implemented in GPyTorch \cite{gardner2018}, while the remaining models are implemented in Scikit-learn \cite{scikit-learn}. We adopt a simple non-personalized baseline which we refer to as \textsc{Mean}. For each person, for each day, we assign the training-data population average  as the number of remaining sedentary periods.


\vspace{-4mm}

\section{Discussion}
\figref{mse} shows that the  two pooling approaches, \gpmt{} and \smallw{} achieve the best overall error rates. 
Furthermore, all of the approaches beat the simple mean prediction baseline. 
However, the extent to which pooling outperforms task-specific models depends on the window length. When longer windows are used in each training period, we see the error rate of the personalized approach decrease, while it increases for pooling. This suggests that waiting longer before retraining can increase the efficacy of
 fully personalized models.
 \figref{delta} shows the time-variance of this data.

As 
users' initial impressions of an intervention's usefulness are crucial, 
 an especially vital question 
 is 
 how well 
the models predict the first few days 
without training data.
There, we see that \gpmt{} achieves the lowest error rates. In these first critical days \gpmt{} achieves a 14\% error reduction over \gpind{}, in the three-day case, and a 9\% reduction in the five-day case. Thus, if predictions must be made early in the study when limited  data is available, some form of pooling might be advantageous.

\section{Conclusion}
Many mobile health intervention treatments are designed to be delivered in a particular context.   Furthermore the number of deliveries per day is often constrained due to concerns about user burden.   Thus, spreading the treatments across the times at which a user is in the desired context is  critical. Here, we propose to personalize the probability of receiving a treatment according to the number of times the context would occur in the remaining part of the day. We consider the context  of being sedentary, and evaluate our ability to predict sedentary periods on a real-world dataset collected from a mobile health study. The data on each individual is small, particularly at the beginning of the study.  Utilizing some form of population-level information can contribute towards rapid personalization for each user, and reduced overall predictive error. We leave many open questions for future work, for example the optimal treatment of the non-stationarity of this data. 

\commentout{
Determining when to send interventions in mobile health is a critical open question. Here, we propose to personalize the probability of receiving an intervention according to the number of times an event might occur. We consider the event of being sedentary, and evaluate our ability to predict sedentary periods on a real-world dataset collected from a mobile health study. This data is small and prone to many issues. Utilizing some form of population level-information can contribute towards overcoming these issues, and reduced overall predictive error. We leave many open questions for future work, for example the optimal treatment of the non-stationarity of this data. }
\subsection*{Acknowledgements}
\vspace{-.2cm}
Research presented in this paper was supported by the
National Heart, Lung and Blood Institute under award number
R01HL125440; the National Institute on Alcohol Abuse and
Alcoholism under award number R01AA023187; the National
Institute on Drug Abuse under award number P50DA039838; and the
National Institute of Biomedical Imaging and Bioengineering under
award number U54EB020404.
\bibliographystyle{plain}
\bibliography{new_bib}

\end{document}